\documentclass{article}
\usepackage[final,nonatbib]{eleg5491_fp}
\usepackage{cite}
\usepackage[utf8]{inputenc} 
\usepackage[T1]{fontenc}    
\usepackage[pdftex,colorlinks]{hyperref} 
\usepackage{url}            
\usepackage{booktabs}       
\usepackage{amsfonts}       
\usepackage{nicefrac}       
\usepackage{microtype}      
\usepackage{graphicx}
\title{Recurrent Soft Attention Model for Common Object Recognition}

\author{
  Liliang Ren \\
  Shanghai Jiao Tong University\\
  \texttt{renll204@sjtu.edu.cn} \\
}

\def\PaperID{1155095838} 

\begin{document}
\maketitle

\begin{abstract}
	We propose the Recurrent Soft Attention Model, which integrates the visual attention from the original image to a LSTM memory cell through a down-sample network. The model recurrently transmits visual attention to the memory cells for glimpse mask generation, which is a more natural way for attention integration and exploitation in general object detection and recognition problem. We test our model under the metric of the top-1 accuracy on the CIFAR-10 dataset. The experiment shows that our down-sample network and feedback mechanism plays an effective role among the whole network structure.

\end{abstract}

\section{Introduction}

The state-of-art convolutional neural networks \cite{yolo,res} have been a great success in a bunch of computer vision tasks. Many researchs \cite{res, he} have been conducted with respect to the structure design of convolutional neural networks for a better feature extraction performance. However, most of the works related to object recognition and objection does not consider the visual attention (i.e. glimpses) which plays an important role in the natural human eyes working process.

\begin{figure}[h]
  \centering
  \includegraphics[scale=0.7]{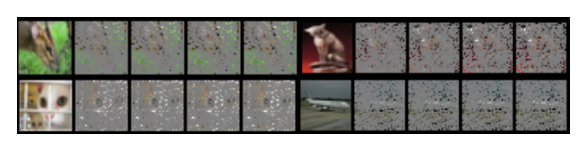}
  \caption{The soft attention masked image generated for each glimpse. The number of bright attention pixels increases as the glimpse number increases, which is in accordance with our daily life experience.}\label{draw}
\end{figure}

In this paper, we propose a recurrent soft attention model which is specialized for common object recognition. We use one classic LSTM cell, propsed by Hochreiter \textit{et al.} \cite{lstm}, for context information memorization and attention mask generation, and the other LSTM cell with the same size for generalizing glimpse information and generating class probabilities. The soft attention is provided by a shallow two-layer convolution network with a downsampling 1x1 convolution layer, and is transfered to the context LSTM cell. The glimpse images generated by the soft attention masks on the original image are shown in Figure \ref{draw}. 

We provided a recurrent soft attention model which apply the visual attention to the common object recognition area. Since the model is quite flexible to be combined with the modern deep convolution neural network (the combination only need replacing the glimpse network), it may provide some inspirations for future computer vision piplines with better performance. The source code can be referenced through the link: \href{https://github.com/renll/RSAM}{\texttt{https://github.com/renll/RSAM}}.

\subsection{Related Work}

Our work is built on some basic researches \cite{ram,mult} with respect to recurrent attention model and reinforcement learning. Mnih \textit{et al.} \cite{ram} uses the reinforcement learning algorithm for training locator, while Ba \textit{et al.}  \cite{mult} uses the Monte Carlo algorithm. However, these works all focus on the recognition of digit numbers which have clear features for attention genralization and classification. Thus, their abilities to recognize common daily-life object are doubted. On the other hand, inspired by the soft attention concept proposed by Xu \textit{et al.} \cite{cap}, we built our soft attention model in a more natural way for attention genralization, which is proved to be effective.

\section{Recurrent Soft Attention Model}
\begin{figure}[h]
  \centering
  \includegraphics[scale=0.55]{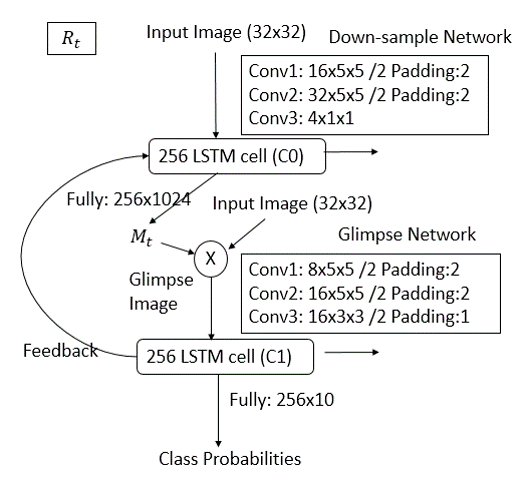}
  \caption{The recurrent soft attention model.}\label{1111}
\end{figure}

Our model is basically a recurrent neural network. The total number of glimpse is a hyperparameter $N$. For each glimpse timestamp $R_{t} (0<t \leq N)$, the input image is down-sampled by a Down-sample Network with two convolution layer each followed by max pooling and a 1x1 convolution layer that down-sampled the feature map numbers to 4. As shown in Figure \ref{1111}.

Then we decode the visual attention from the hidden units in the LSTM cell through a fully-connected layer with the ReLu activation function, and the obtained mask matrix $M_t$ from the LSTM cell (C0) is used for generating the glimpse image by multiplying the input image pixel matrix element-wisely. This type of multiplicative interaction between the features and the location was initially proposed by Larochelle \textit{et al.}\cite{comb}. Since the ReLu activation function is used, the mask matrix will contain many zero elements and some non-zero elements encoding the attenetion information, which largely reduces the required computation resources needed for the following glimpse network. This is exactly what we want for a visual atttention model.

In the end of a glimpse timestamp, the hidden units of the LSTM cell (C1) is inputted to the LSTM cell (C0) as a feedback to provide the glimpse information needed for the memory units inside C0 to conduct the next attention extraction process. In this way, the attention generating process will not only be based on the context information, but also depend on the result from the glimpse process, which acts as a simplified feedback mechanism during the natural visual attention process proposed by G. Deco \textit{et al.}\cite{fed}. We minimize the cross entropy loss between the output and the target in each glimpse timestamp. The ReLu nonlinear activation function is used throughout the network.

\section{Experiments}

We use CIFAR-10, which consists of 50 thousanfd training images and 10 thousand testing images in 10 classes, as our dataset for testing. We conduct experiments that are trained on the training set with the mini-batch size of 128 and evaluated on the test set. Our focus is on the behaviors of our down-sample network and the feedback mechanism, but not on pushing the state-of-the-art results. That is the reason why we intentionally use rather less neurons in each network layer and scale down the whole network.

\begin{figure}[h]
  \centering
  \includegraphics[scale=0.7]{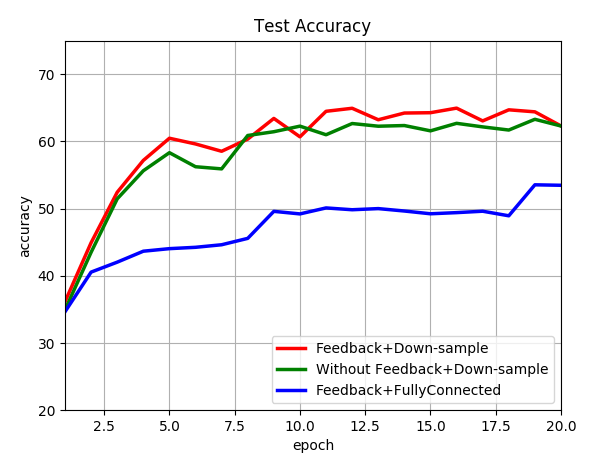}
  \caption{The comparison of the contribution to the Top-1 test accuracy between each component. (The down-sample network is replaced by a fully-connected layer proposed by Xu \textit{et al.} \cite{cap} when introducing the soft attention concept)}\label{222}
\end{figure}

\begin{figure}[h]
  \centering
  \includegraphics[scale=0.7]{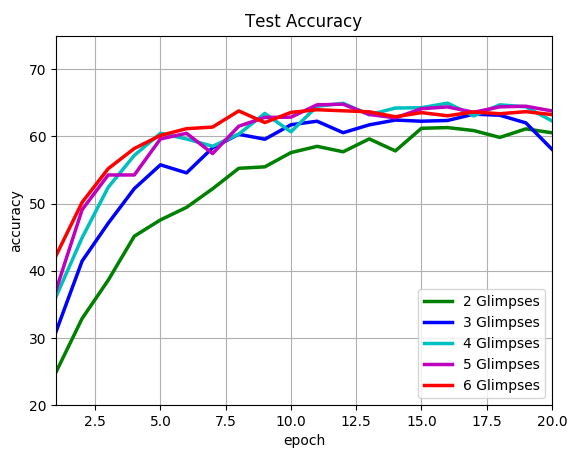}
  \caption{The Top-1 test accuracy comparison between different glimpse numbers.}\label{LOL}
\end{figure}

The model is trained under the circumstance of 4 glimpses, using the stochastic gradient decent method with a momentum of 0.9 and the weight decay of 0.0001. The initial learning rate is 0.01 with an exponentially drop of 0.95 after each epoch. We conduct our training on a g2.x2large instance on the Amazon Web Service. The whole model does not implment any regularization or data augmentation tricks, and only batch normalization is added to each neuron layer (except the last classification layer) for stablizing the weight signal transmission. The final output probability is an average of the softmax probabilities from each of the glimpse timestamp.

From the Figure \ref{222}, we can see that both the feedback mechanism and the down-sample network contributes to the test accuracy of the model. The feedback mechanism boosts the model learning pace and leads to more robust covergence, while the down-sample network provide the effective attention information for location. The best performance is achieved by the model with both of the components.

We also test our model with different glimpse numbers for 20 epochs, under the same model setting of feedback plus down-sample network. From the Figure \ref{LOL}, we can see that as the glimpse number increases, the accuracy will converge as our expectation that the total information provided by a single image is limited if the feature extractor is not improved. The highest test accuracy 64.95\% is achieved by the model with 4 glimpses.

\section*{Acknowledgments}

We would like to thank Tong Xiao, Xiaogang Wang from the Chinese University of Hong Kong for many helpful comments and discussions.

\bibliographystyle{ieee.bst}
\bibliography{eleg5491.bib}

\begin{thebibliography}{1}\itemsep=-1pt

\bibitem{mult}
J.~Ba, V.~Mnih, and K.~Kavukcuoglu.
\newblock Multiple object recognition with visual attention.
\newblock {\em arXiv preprint arXiv:1412.7755}, 2014.

\bibitem{fed}
G.~Deco and J.~Zihl.
\newblock A neurodynamical model of visual attention: feedback enhancement of
  spatial resolution in a hierarchical system.
\newblock {\em Journal of computational neuroscience}, 10(3):231--253, 2001.

\bibitem{he}
K.~He, X.~Zhang, S.~Ren, and J.~Sun.
\newblock Delving deep into rectifiers: Surpassing human-level performance on
  imagenet classification.
\newblock In {\em Proceedings of the IEEE international conference on computer
  vision}, pages 1026--1034, 2015.

\bibitem{res}
K.~He, X.~Zhang, S.~Ren, and J.~Sun.
\newblock Deep residual learning for image recognition.
\newblock In {\em Proceedings of the IEEE Conference on Computer Vision and
  Pattern Recognition}, pages 770--778, 2016.

\bibitem{lstm}
S.~Hochreiter and J.~Schmidhuber.
\newblock Long short-term memory.
\newblock {\em Neural computation}, 9(8):1735--1780, 1997.

\bibitem{comb}
H.~Larochelle and G.~E. Hinton.
\newblock Learning to combine foveal glimpses with a third-order boltzmann
  machine.
\newblock In {\em Advances in neural information processing systems}, pages
  1243--1251, 2010.

\bibitem{ram}
V.~Mnih, N.~Heess, A.~Graves, et~al.
\newblock Recurrent models of visual attention.
\newblock In {\em Advances in neural information processing systems}, pages
  2204--2212, 2014.

\bibitem{yolo}
J.~Redmon, S.~Divvala, R.~Girshick, and A.~Farhadi.
\newblock You only look once: Unified, real-time object detection.
\newblock In {\em Proceedings of the IEEE Conference on Computer Vision and
  Pattern Recognition}, pages 779--788, 2016.

\bibitem{cap}
K.~Xu, J.~Ba, R.~Kiros, K.~Cho, A.~Courville, R.~Salakhudinov, R.~Zemel, and
  Y.~Bengio.
\newblock Show, attend and tell: Neural image caption generation with visual
  attention.
\newblock In {\em International Conference on Machine Learning}, pages
  2048--2057, 2015.

\end{thebibliography}

\end{document}